# Global 6DOF Pose Estimation from Untextured 2D City Models

Clemens Arth, Christian Pirchheim, Vincent Lepetit
Graz University of Technology
{arth,pirchheim,lepetit}@icg.tugraz.at

Jonathan Ventura
University of Colorado Colorado Springs
jventura@uccs.edu

## Abstract

*We propose a method for estimating the 3D pose for the camera of a mobile device in outdoor conditions, using only an untextured 2D model. Previous methods compute only a relative pose using a SLAM algorithm, or require many registered images, which are cumbersome to acquire. By contrast, our method returns an accurate, absolute camera pose in an absolute referential using simple 2D+height maps, which are broadly available, to refine a first estimate of the pose provided by the device's sensors. We show how to first estimate the camera absolute orientation from straight line segments, and then how to estimate the translation by aligning the 2D map with a semantic segmentation of the input image. We demonstrate the robustness and accuracy of our approach on a challenging dataset.*

## 1. Introduction

While tremendous progress has been made over the last years in camera registration by combining image information and pose and motion sensors, current methods still do not allow large-scale image-based 3D localization. Some methods compute the pose from matches between the input images and pre-registered images [19, 18]. However, creating the registered images in the first place is still time-consuming, and requires many images, which are often available only for popular places. Other methods, based on Simultaneous Localization and Mapping (SLAM) use only the input images but are restricted to provide "relative" poses, meaning that the poses are recovered only up to a rigid motion. This is not enough for many applications, such as navigation aid [11] or labeling of local touristic landmarks [25].

As shown in Fig. 1, we therefore introduce a method that relies on untextured 2D maps augmented with an approximative height of the buildings to provide an absolute and accurate pose that can be used for accurate 3D localization. Such 2D maps are widely available in practice and can be obtained for example from OpenStreetMap [1], which offers maps with a very broad coverage, and free to use under an open license.

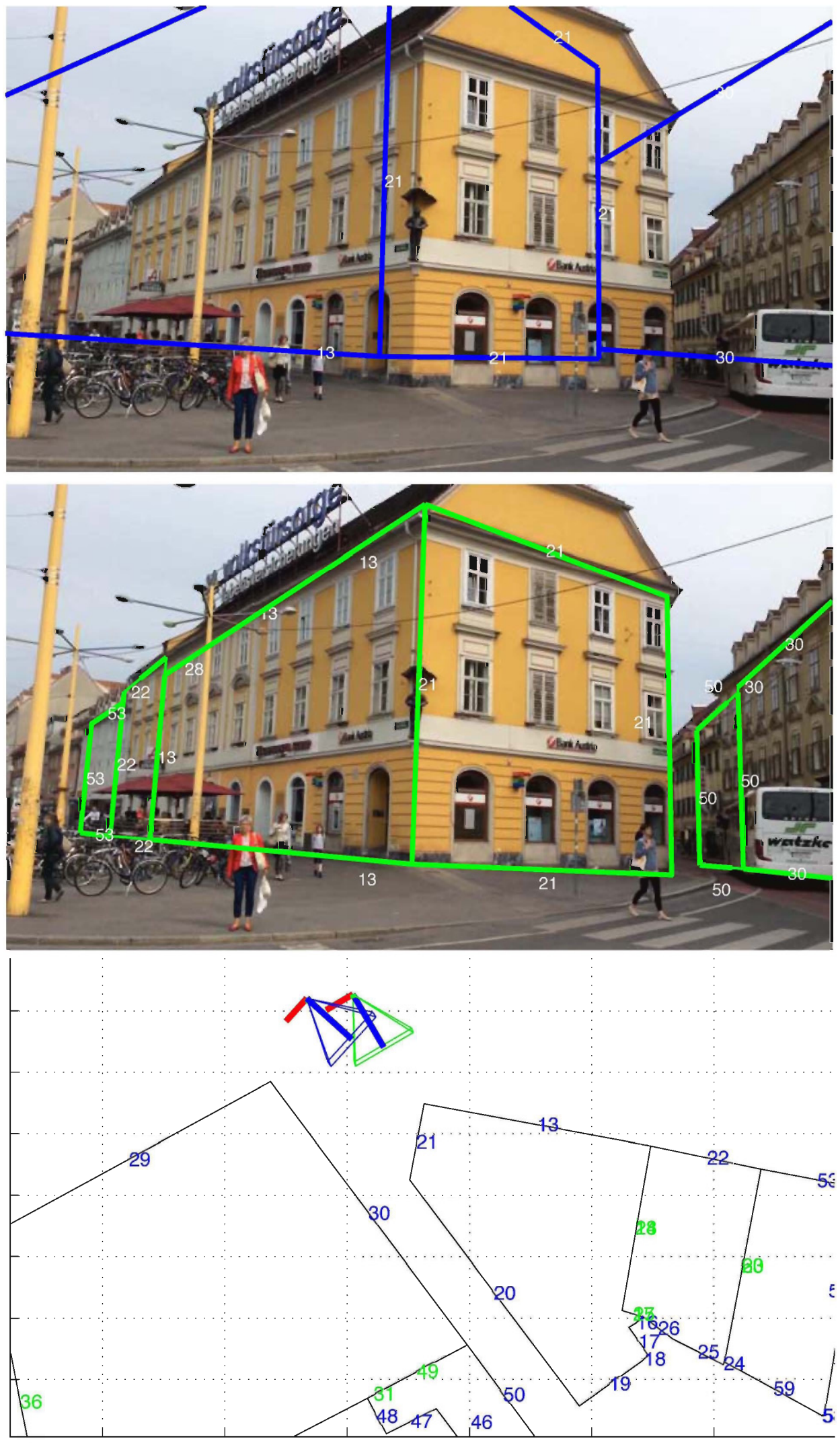

Figure 1: **Top:** Model reprojection into the pose estimate provided by the sensors. **Middle:** Model reprojection after correction by our method. **Bottom:** Our method relies only on a 2D map and the heights of the buildings, in addition to the input image itself.

To the best of our knowledge, this is a novel approach to outdoor 3D registration, but also a very challenging one, as untextured 2D maps bring very little information. We also exploit the pose estimate provided by the sensors, but it can be very noisy as the one shown in Fig. 1.

In order to estimate the *absolute* orientation of the camera, we assume that most of the straight line segments that can be extracted from the buildings' façades are either horizontal or vertical. This is a common assumption used in vanishing point and relative orientation estimation. It is valid for many urban environments, and tolerates windows with relatively complex shapes as the ones of the buildings in our datasets.

This assumption together with the 2D maps allows us to estimate the *absolute* orientation of the camera, but it does not constrain the camera translation. We therefore propose to solve for the translation by aligning the buildings model with a semantic segmentation of the image.

Maybe the closest methods to our approach are the works by Ramalingam *et al*. [15] and Chu *et al*. [3]. The former is based on 3D models, SIFT features, and edges. Correspondences between multiple images are used to estimate accurate poses and only a qualitative assessment is given. The latter work proposes an approach to improve GPS estimates using descriptors of building corners formed by edges and vanishing directions. While the results are encouraging, however, the accuracy in terms of orientation and position is far from sufficient for Augmented Reality applications for example. By contrast, our approach provides a highly accurate 6DOF pose using a single image.

In the remainder of the paper, we first discuss related work in more details, and then describe our method, finally evaluating it on a dataset acquired with a state-of-the-art mobile device.

## 2. Related Work

Many approaches to outdoor registration have been developed, and we give here only a summary.

The first proposed systems were based on sensors only [6]. However it quickly became clear that sensors are not accurate enough by themselves for precise 3D augmentations, and combining sensors with image information became a natural approach: Digital Elevation Model (DEM) data was used in [10] together with Shape Context descriptors for horizon recognition. Later, [16] described a model-based tracking system in urban environments, combining sensors, textured 3D models, and edgels. [19] then showed location recognition was possible using images only by storing millions of features from 20 km of urban street-side imagery taken from a moving vehicle, organized in a vocabulary tree to handle the massive amounts of data. Since then, many works based on image retrieval have been proposed [8, 28, 9, 18, 1].

These methods rely on a single image, but a few others exploit several images. While, as pointed out in the introduction, SLAM systems can only provide relative poses, they are still useful to compute absolute poses: Recently, [14] combined a real-time SLAM system on mobile devices, together with a globally registered SFM reconstruction, providing highly accurate 6DOF tracking by aligning a local SLAM map globally over time.

Image-based registration approaches rely on pre-captured images that were registered offline, which is time-consuming and does not scale well to world-wide areas. [26] showed it was possible to use Google Street View images to get an estimate of the camera pose. Later, [22] improved upon this, estimating the geospatial trajectory of a camera with unknown intrinsic parameters in urban environments. However, Google Street View images are not available for every country, but much more critically it is still very challenging to match images under arbitrary lighting in outdoor conditions, and Google Street View provides only scarcely sampled images.

By contrast, our approach relies on broadly available data, making it much more practical, and does not rely on feature point matching but rather on straight line segments, which are much more stable to lighting variations.

## 3. 2D Map-Based Pose Estimation from a Single Image

Our approach proceeds is two steps. Using an initial pose estimation from the device's sensors for the input image, we retrieve the 2D model of the surrounding buildings, and estimate first the camera orientation in a absolute coordinate system. Given this orientation, we then estimate the translation.

### 3.1. Orientation Estimation

We start by computing the pitch and roll of the camera, *i.e*. orientation of the camera's vertical axis with respect to gravity, from line segments. This can be performed without using any information from the 2D map: It is a relatively standard problem in the vanishing point literature, and we describe briefly our approach. The estimation from such data of the yaw, the last degree-of-freedom of the rotation, in the absolute referential is, to the best of our knowledge, a new problem.

[1] http://www.openstreetmap.org

### 3.1.1 Estimating the Camera's Vertical Axis

We want to estimate a rotation matrix $\mathbf{R}_v$ that aligns the camera's vertical axis with the gravity vector. We do so by determining the dominant vertical vanishing point in the image, using line segments extracted from the image. We rely on the Line Segment Detector (LSD) algorithm [24], followed by three filtering steps: we only retain line segments exceeding a certain length; lines below the horizon line computed from the rotation estimate of the sensor are removed, since these segments are likely located on the ground plane or foreground object clutter; line segments are removed if the angle between their projection and the gravity vector given by the sensor is larger than a threshold [13]. The intersection point $\mathbf{p}$ of the projections $\mathbf{l}_1$ and $\mathbf{l}_2$ of two vertical lines is the vertical vanishing point, and can be computed with as a cross product:

$$\mathbf{p} = \mathbf{l}_1 \times \mathbf{l}_2 \tag{1}$$

using homogeneous coordinates. As suggested in [17], we use exhaustive search of all pairs of lines in order to find the dominant vanishing point. For each pair of vertical line segments, we compute the intersection point and test it against all line segments, using an angular error measure:

$$\text{err}(\mathbf{p},\mathbf{l}) = \text{acos}\left(\frac{\mathbf{p}\cdot\mathbf{l}}{||\mathbf{p}||\cdot||\mathbf{l}||}\right). \tag{2}$$

The dominant vertical vanishing point $\mathbf{p}_v$ is chosen as the one with the highest number of inliers, using an error threshold of $\sigma$ degrees.

Given the dominant vertical vanishing point $\mathbf{p}_v$, we now compute the rotation which would align the camera's vertical axis with the vertical vanishing point. The vertical direction is $\mathbf{z} = [0\ 0\ 1]^\top$. Using angle-axis representation, the axis of the rotation is $\mathbf{u} = \mathbf{p}_v \times \mathbf{z}$ and the angle is $\theta = \text{acos}(\mathbf{p}_v \cdot \mathbf{z})$, assuming that the vertical vanishing point is normalized. The rotation $\mathbf{R}_v$ then can be constructed using $SO(3)$ exponentiation:

$$\mathbf{R}_v = \exp_{SO(3)}\left(\mathbf{u}\cdot\frac{\theta}{||\mathbf{u}||}\right). \tag{3}$$

At this point, the camera orientation is determined up to a rotation around its vertical axis (yaw). We explain in the next section how to estimate this last degree-of-freedom for the orientation.

### 3.1.2 Orientation in the Absolute Coordinate System

Fig. 2 gives our key idea to estimate the camera rotation around the vertical axis in the absolute coordinate system: We look for the rotation that makes the line segments corresponding to horizontal 3D edges, actually horizontal when they are back-projected on a simple 3D model inferred from the 2D map.

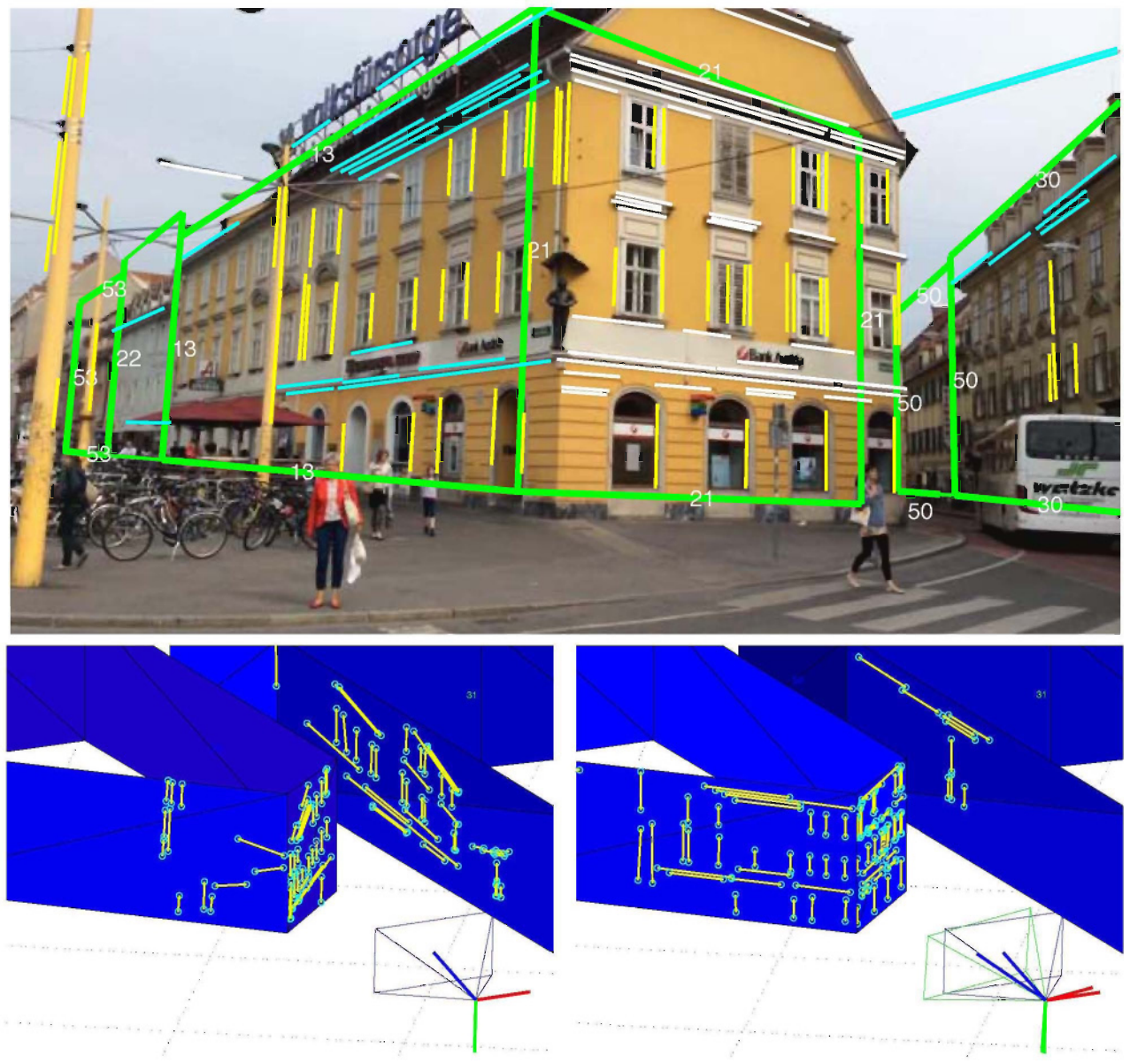

Figure 2: Estimating rotation $\mathbf{R}_h$. **Top:** line segments identified as vertical (yellow) and horizontal (cyan) in 3D, and the reprojection of the model after rotation correction (green), but before translation correction. **Bottom left:** back-projection of the line segments using the sensor pose. We estimate the rotation around the vertical axis by making the back-projection of the cyan segments horizontal. **Bottom right:** reprojection of the line segments after rotation correction.

Given a façade $f$ from the 2D map, its horizontal vanishing point is found as the cross product of its normal $\mathbf{n}_f$ and the vertical axis $\mathbf{z}$:

$$\mathbf{p}_h = \mathbf{n}_f \times \mathbf{z}. \tag{4}$$

After orientation correction through $\mathbf{R}_v$, the projection of horizontal lines lying on $f$ should intersect $\mathbf{p}_h$. Thus, given a horizontal vanishing point $\mathbf{p}_h$ and the projection of a horizontal line segment $\mathbf{l}_3$, we can compute the rotation $\mathbf{R}_h$ about the vertical axis, to align the camera's horizontal axis with the horizontal vanishing point of $f$. This rotation has one degree of freedom, $\phi_z$, the amount of rotation about the vertical axis:

$$\mathbf{R}_h = \begin{bmatrix} \cos\phi_z & -\sin\phi_z & 0 \\ \sin\phi_z & \cos\phi_z & 0 \\ 0 & 0 & 1 \end{bmatrix}. \tag{5}$$

Using the substitution $q = \tan\frac{\phi_y}{2}$ we get $\cos\phi_z = \frac{1-q^2}{1+q^2}$ and $\sin\phi_z = \frac{2q}{1+q^2}$ [12]. We can therefore parameterize our rota-tion matrix in terms of $q$:

$$\mathbf{R}_h = \frac{1}{1+q^2}\begin{bmatrix} 1-q^2 & -2q & 0 \\ 2q & 1-q^2 & 0 \\ 0 & 0 & 1+q^2 \end{bmatrix}. \quad (6)$$

The intersection constraint between $\mathbf{l}_3$ and the horizontal vanishing point $\mathbf{p}_h$ is expressed as

$$\mathbf{p}_h \cdot (\mathbf{R}_h \mathbf{l}) = 0\,, \quad (7)$$

giving a single quadratic polynomial in $q$. The roots of the polynomial determine two possible rotations. This ambiguity is resolved by choosing the rotation which makes the camera's view vector opposite that of the normal $\mathbf{n}_f$.

In practice we create pairs $< \mathbf{l}, f >$ from line segments $\mathbf{l}$ assigned to visible façades $f$, identified from the 2D map using the initial pose estimate provided by the sensors, and using a Binary Space Partition (BSP) tree [7] for efficient assignment. We exhaustively evaluate the angular error measure from Eq. (2) for a rotation estimate from the pair $< \mathbf{l}, f >$ for all remaining pairs, choosing the hypothesis with the highest number of inliers.

Finally, the absolute rotation $\mathbf{R}$ of the camera is computed by chaining the two previous rotations $\mathbf{R}_v$ and $\mathbf{R}_h$, *i.e.*

$$\mathbf{R} = \mathbf{R}_v \mathbf{R}_h\,. \quad (8)$$

Fig. 2 gives an example of this orientation estimation procedure, showing the line segments used for this estimation.

## 3.2. Translation Estimation

While the vertical and horizontal segments lying on the façades allow us to estimate the camera's orientation in a global coordinate frame, it is easy to realize that horizontal and vertical segments within the façades do not provide any useful constraint to estimate the translation since we do not know their exact 3D location. The pose could in theory be computed from correspondences between the edges of the buildings in the 2D map, and their reprojections in the images. In practice, it is virtually impossible to directly obtain such matches reliably in absence of additional information.

Our key idea to estimate the translation is to align the 2D map with a semantic segmentation of the image: We can estimate the translation of the camera as the one that aligns the façades of the 2D map with the façades extracted from the image.

To speed up this alignment, and to make it more reliable, we first generate a small set of possible translations given the line segments in the image that potentially correspond to the edges of the buildings in the 2D map. We then keep the hypothesis that aligns the 2D map with the segmentation the best. We detail these two steps below.

### 3.2.1 Generating Translation Hypotheses

In practice, the translation along the vertical axis is the most problematic one to estimate from the image, because the bottoms of the buildings are typically occluded by pedestrians. We therefore simply set the height of the camera at 1.6 $m$, which is reasonable for a handheld device.

We generate possible horizontal translations for the camera by matching the edges of the buildings with the image. However, this is a very challenging task, as the images are very cluttered in practice.

As shown in Fig. 3, we generate a set of possible image locations for the edges of the buildings with a simple heuristic. We first rectify the input image using the orientation so that vertical 3D lines also appear vertical in the image, and we sum the image gradients along each column. The columns with a large sum are likely to correspond to the border of a building. However, since windows also have strong vertical edges, they tend to generate many wrong hypotheses. To reduce their influence, we trained a multi-scale window detector [23]. Pixels lying on the windows found by the detector are ignored when computing the gradient sums over the columns. We also use the façade segmentation result described in Section 3.2.2 to consider only the pixels that lie on façades, but not on windows. Since the sums may take very different values for different scenes, we use a threshold estimated automatically for each image: We fit a Gamma distribution to the histogram of the sums and evaluate the quantile function with a fixed inlier probability.

Finally, as shown in Fig. 3(g) and Fig. 3(h), we generate translation hypotheses for each possible pair of correspondences between the vertical lines extracted from the image and the building corners. The building corners come from the corners in the 2D maps that are likely to be visible, given the location provided by the GPS and the orientation estimated during the first step. Given two vertical lines in the image, $\mathbf{l}_1$ and $\mathbf{l}_2$, and two 3D points which are the corresponding building corners, $\mathbf{x}_1$ and $\mathbf{x}_2$, the camera translation $\mathbf{t}$ in the ground plane can be easily computed by solving the following linear system:

$$\begin{cases} \mathbf{l}_1 \cdot (\mathbf{x}_1 + \mathbf{t}) & = 0 \\ \mathbf{l}_2 \cdot (\mathbf{x}_2 + \mathbf{t}) & = 0 \end{cases}. \quad (9)$$

### 3.2.2 Aligning the 2D Map with the Image

To select the best translation among the ones generated using the method described above, we evaluate the alignments of the image and the 2D map after projection using each generated translations.

We use a very simple pixel-wise segmentation of the input image, by applying a classifier to each image patch of a given size to assign a class label to the center location of the patch. Much more sophisticated methods could be used [21, 5], however we can still estimate reliably the camera translation despite the relatively poor quality of our segmentation.

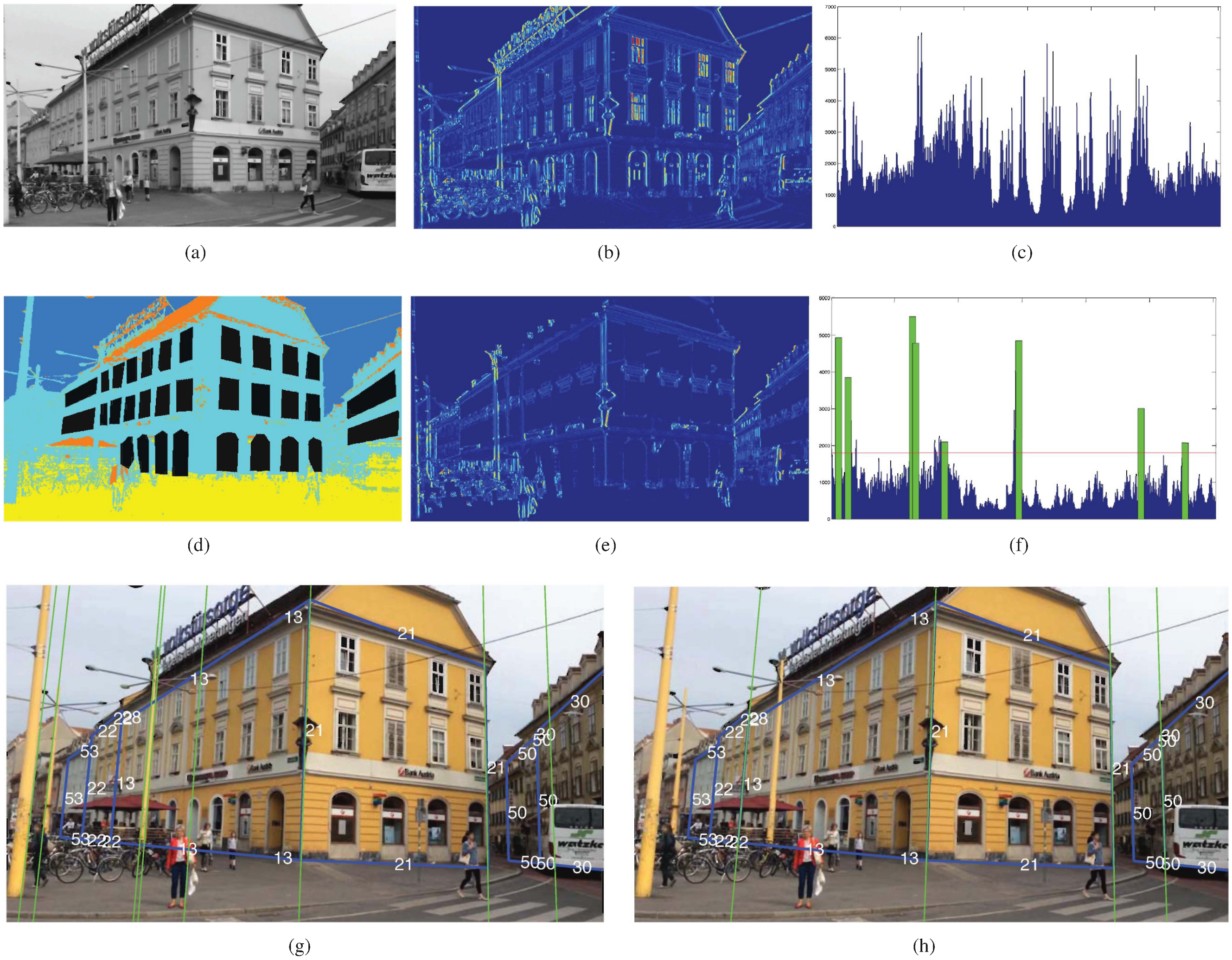

Figure 3: Generating translation hypotheses. **(a)**: Vertically rectified image. **(b)**: Image gradients. **(c)**: Histogram of the sums of the gradient magnitude over the columns. **(d)**: Segmentation of the façades in cyan and window detections in black. **(e)**: Image gradients only for the pixels lying on a façade but not on a window. **(f)**: Histogram of gradient sums, and selected vertical image lines. **(g)**: Selected image lines overlayed on the original image. **(h)**: 3D model lines from building corners overlayed on the original image using the ground truth pose, most of them were detected with our method.

More exactly, we apply a multi-class Support Vector Machine (SVM) [20, 2] we trained on a dataset of manually segmented images, different from the images used to evaluate our registration method. We use the integral features introduced in [4], and consider five different classes $C = \{c_f, c_s, c_r, c_v, c_g\}$ for *façade, sky, roof, vegetation* and *ground*, respectively. By applying the classifier exhaustively, we obtain a probability estimate $p$ for each image pixel over these classes. Fig. 4 shows an example of a segmentation for a typical input image.

As illustrated in Fig. 5, given the 2D projection $\text{Proj}(M, \mathbf{p})$ of our 2D map+height $M$ into the image using pose hypothesis $\mathbf{p}$, we compute the log-likelihood of the pose:

$$s_{\mathbf{p}} = \sum_{i}^{\text{Proj}(M,\,\mathbf{p})} \log p_i(c_f) + \sum_{i}^{\neg\text{Proj}(M,\,\mathbf{p})} \log\left(1 - p_i(c_f)\right) \,, \quad (10)$$

where $\neg\text{Proj}(M, \mathbf{p})$ denotes the set of pixels lying outside the reprojection $\text{Proj}(M, \mathbf{p})$. The pixels lying on the projection $\text{Proj}(M, \mathbf{p})$ of the façades should have a high probability to be on a façade in the image, and the pixels lying outside should have a high probability to not be on a façade.

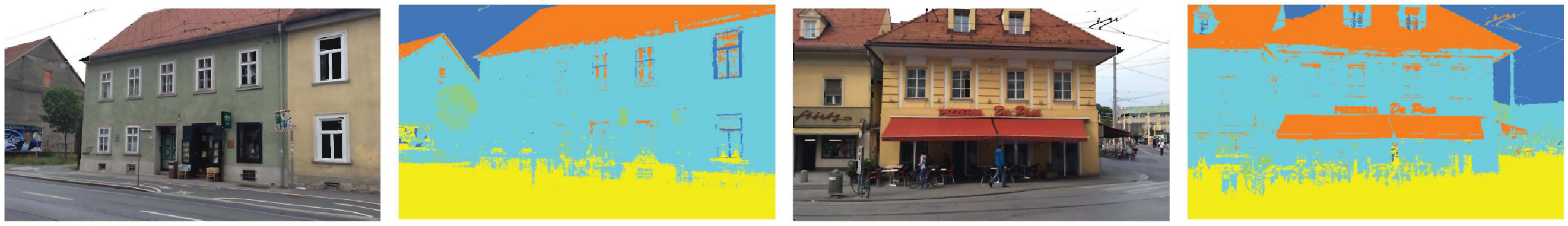

Figure 4: Pixel-wise segmentations we obtain with a simple multi-class SVM for two different images. Cyan corresponds to façades, blue to sky, orange to roofs, green to vegetation, and yellow to ground plane.

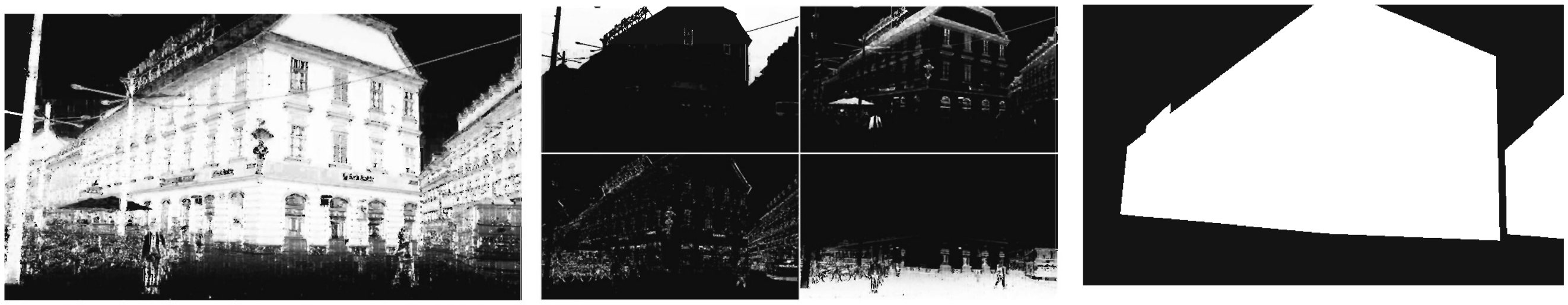

Figure 5: Computing the log-likelihood. **Left:** Probability map for $c_f$, the façade class. **Middle:** Probability maps for $c_s, c_r, c_v$ and $c_g$. **Right:** Reprojection $\text{Proj}(M,\ \mathbf{p})$ for a pose close to the ground truth.

We keep the pose $\hat{\mathbf{p}}$ that maximizes the log-likelihood:

$$\hat{\mathbf{p}} = \arg\max_{\mathbf{p}} s_{\mathbf{p}} \,. \tag{11}$$

In practice, the 3D pose estimated from the sensors is often not accurate enough to directly initialize our method. We therefore sample six additional poses around this pose, along a circle with a 12.5 *m* radius [27], execute our method initialized from each of these seven poses, and keep the computed pose with the largest likelihood.

Note that this approach naturally extends to more complex building models, for example if the roofs of the buildings are also present in the model. The log-likelihood then becomes:

$$s_{\mathbf{p}} = \sum_{c \in C_M} \sum_{i}^{\text{Proj}(M_c,\ \mathbf{p})} \log p_i(c) + \sum_{i}^{\neg \text{Proj}(M,\ \mathbf{p})} \log \left( 1 - \sum_{c \in C_M} p_i(c) \right) , \tag{12}$$

where $C_M$ is a subset of $C$ and made of the different classes that can appear in the buildings model, and $\text{Proj}(M_c,\ \mathbf{p})$ the projection of the components of the buildings model for class $c$.

# 4. Experimental Results

In this section, we first describe the dataset we built to evaluate our approach, and then report and discuss the results of the evaluation.

## 4.1. Dataset

To demonstrate the applicability of our approach, we captured a dataset of 32 images with an *Apple iPad Air* in urban and suburban environments from a European city [2]. The images were captured without any special consideration for satellite shadowing nor surrounding metallic structures.

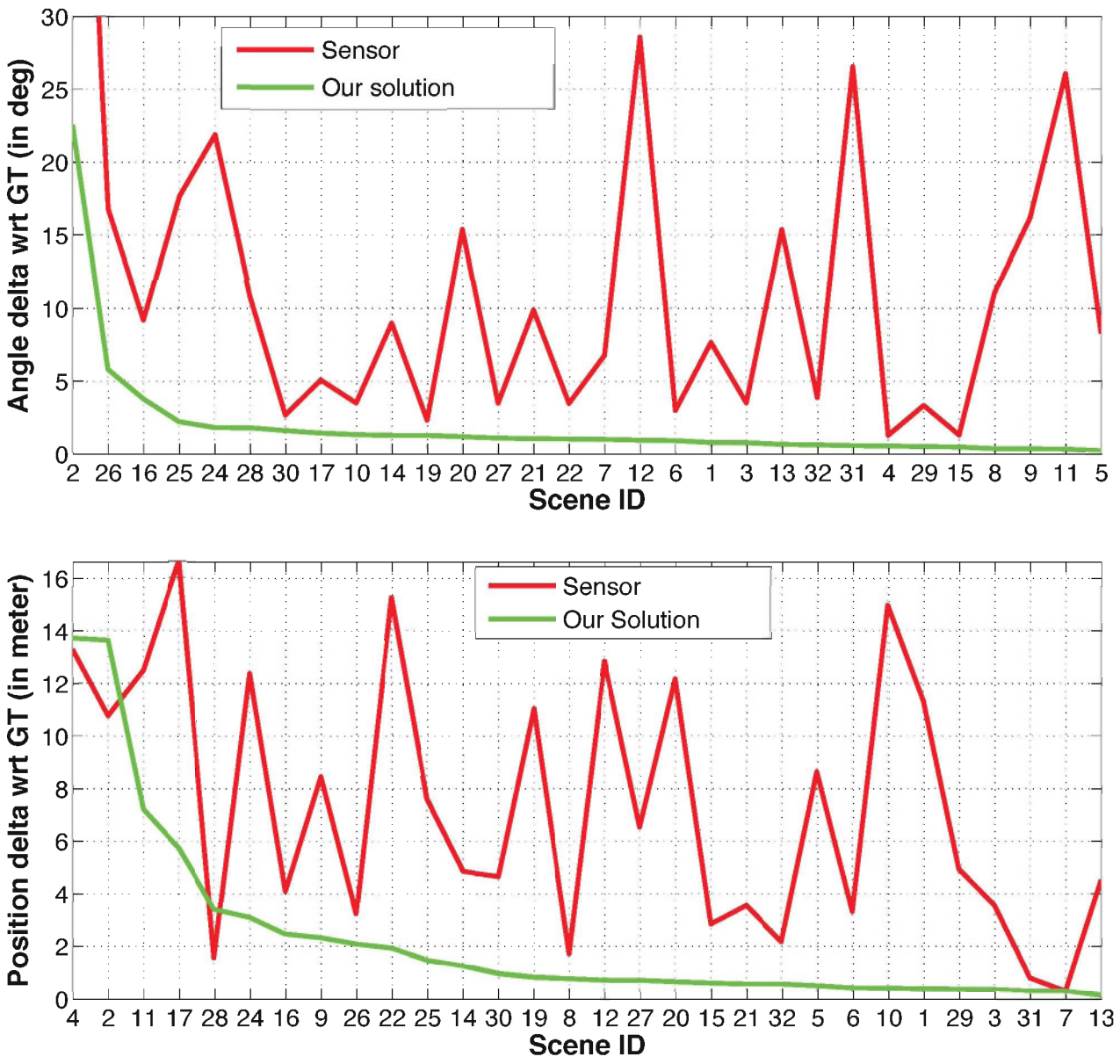

Figure 6: Pose estimates accuracy. **Top:** Orientation, and **Bottom:** Translation. We ranked the images from the one with the largest error after correction to the one with the smallest error. Our method significantly improves the accuracy of the orientation and translation estimates.

[2] We will make our dataset including our 32 images, the sensor poses, the ground truth poses, and the 2D+height maps publicly available

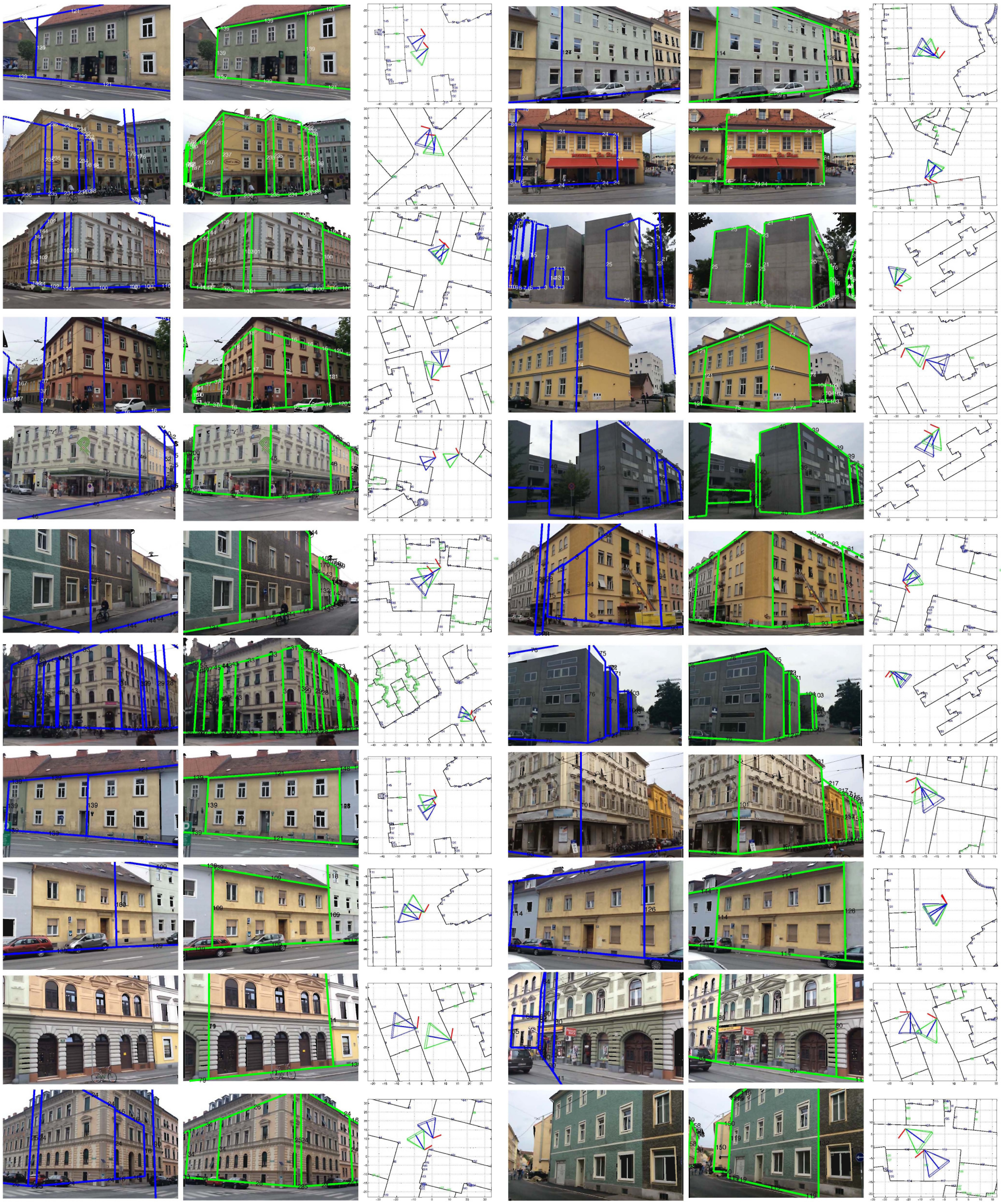

Figure 7: Results of our approach on test images. For each triplet of images: **Left:** Model reprojection into the image using the initial sensor pose. **Middle:** Model reprojection into the image using the final estimated pose. **Right:** Map view, sensor pose (in blue) and the corrected pose (in green).

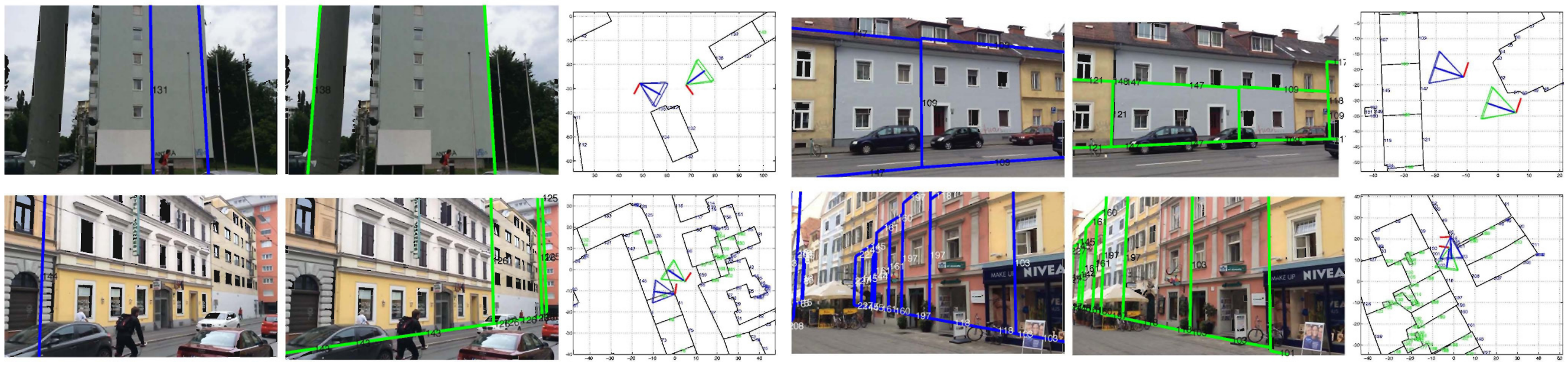

Figure 8: Images with the largest pose errors. For each triplet of images: **Left:** Model reprojection into the image using the initial sensor pose. **Middle:** Model reprojection into the image using the final estimated pose. **Right:** Map view, sensor pose (in blue) and the corrected pose (in green). Even for these images, the model reprojection tends to be close to the expected position.

As a consequence, the accuracy of the pose estimated with the sensors only ranges from very accurate, about 0.4 *m* position and 2° rotation error, to very poor, up to 16.5 *m* position and about 30° rotation error. Since altitude estimates from sensors tend to be very poor, we reset the altitude estimates of the poses predicted by the sensors to a default value of 1.6 *m*. For each test image, we calculated a ground truth pose by manually matching 2D image locations with 3D points from the maps.

We retrieved 2D maps of the surroundings from OpenStreetMap and extended them with a coarse estimate of the height of the buildings, obtained from aerial laser scan height data provided by local authorities.

### 4.2. Orientation Accuracy

Fig. 6-top plots the angular error of the camera pose predicted by the sensors, and after correction with our method. The error is calculated as the angle difference between the estimated rotation and the ground-truth rotation in angle-axis representation. We ranked the images from the one with the largest error after correction to the one with the smallest error. The sensors make errors that can reach up to 30°. After our method, all our orientation estimates have an angle error below 5°, with the exception of a single outlier image which contains very little horizontal lines. 90.6% of the estimates are below 3°, 84.4% below 2° and 50% below 1° of angular error with respect to the ground truth rotation.

### 4.3. Translation Accuracy

Fig. 6-bottom gives the results of our translation estimation method. As for the rotation, we ranked the images from the one with the largest error after correction to the one with the smallest error. The sensor errors range from about 0.4 *m* to about 16.5 *m*, with an average error of about 8 *m*. Our method significantly decreases the translation error in most of the cases. The worst results are due to buildings next to each other, with edges that cannot be extracted correctly. Overall, our method is able to considerably improve the position estimates from the sensors, with the pose estimates for 87.5% of the images being below 4 *m*, 68.8% below 2 *m* and 59.4% below 1 *m* of error with respect to the ground truth position.

### 4.4. Visual Inspection

Fig. 7 presents the final results of our algorithm for many different images, showing the reprojection of the model using both the sensor pose and the pose retrieved from our approach. After pose estimation, the outlines of the models nicely fit the building outlines, even for very challenging scenes with many façades visible and a considerable rotation and position error in the sensor estimate. The amount of correction can be assessed from the map view, as both the rotation and translation undergo a significant correction during the application of our method.

Fig. 8 shows the images with the largest pose errors. Failures are mostly due to buildings next to each other, without clear separation between them.

## 5. Conclusion

We presented a novel approach to accurate 6DOF pose estimation that relies only on 2D maps. Our approach takes sensor estimates and corrects the rotation estimate, followed by fitting an untextured 3D model to an automated segmentation of the input image. Our approach is evaluated on a challenging dataset of real-world images and sensor data, and considerably improved the pose estimates both in terms of rotation and translation.

Our approach can be used to initialize a SLAM algorithm, which will thus provide the camera pose in the coordinate system of the 2D maps. We plan to use this approach in Augmented Reality applications on mobile devices.

While our method already provides good results, it can certainly be improved. We hope that our work and dataset will encourage new research on the problem we introduce in this paper.